\documentclass[a4paper, 10pt, conference]{ieeeconf}

\IEEEoverridecommandlockouts                       

\usepackage{graphics} 
\usepackage{epsfig} 
\usepackage{mathptmx} 
\usepackage{times} 
\usepackage{amsmath} 
\usepackage{amssymb}  

\usepackage{lineno}
\usepackage{tikzpagenodes}
\usepackage{background}
\usepackage{gensymb}
\usepackage{subcaption}
\usepackage{tensor}
\usepackage{hyperref}
\usepackage{booktabs}
\usepackage[binary-units=true]{siunitx}

\usepackage[utf8]{inputenc}

\usepackage{pgfplots}
\usepackage{tikz-3dplot}

\usetikzlibrary{calc}
\usetikzlibrary{arrows}
\usetikzlibrary{angles}
\usetikzlibrary{quotes}
\usetikzlibrary{decorations.pathreplacing}
\usetikzlibrary{bending}
\usetikzlibrary{calc}

\backgroundsetup{color=white}

\usepackage[format=plain,
            labelfont={bf,it},
            textfont=it]{caption}

\title{\LARGE \bf 3D Robot Pose Estimation from 2D Images}

\author{Christoph Heindl$^{1}$, Sebastian Zambal$^{1}$, Thomas Pönitz$^{1}$, Andreas Pichler$^{1}$, and Josef Scharinger$^{2}$
\thanks{$^{1}$Profactor GmbH, Im Stadtgut A2, 4407 Steyr-Gleink, Austria}%
\thanks{$^{2}$JKU Department of Computational Perception, Altenbergerstr. 69, 4040 Linz, Austria}%
}
\begin{document}

\maketitle

\begin{abstract}
    This paper considers the task of locating articulated poses of multiple robots in images. Our approach simultaneously infers the number of robots in a scene, identifies joint locations and estimates sparse depth maps around joint locations. The proposed method applies staged convolutional feature detectors to 2D image inputs and computes robot instance masks using a recurrent network architecture. In addition, regression maps of most likely joint locations in pixel coordinates together with depth information are computed. Compositing 3D robot joint kinematics is accomplished by applying masks to joint readout maps. Our end-to-end formulation is in contrast to previous work in which the composition of robot joints into kinematics is performed in a separate post-processing step. Despite the fact that our models are trained on artificial data, we demonstrate generalizability to real world images.
\end{abstract}

\section{INTRODUCTION}
    
    \label{sec:intro} 
    In this work we consider the task of estimating kinematic robot chains in images and videos. This technology enables enriched scene understanding for mobile robotics in unknown industrial environments. In particular the task of spatio-temporal coordination with static or other mobile robotic entities is avoided. While the estimation of human and animal body parts in images has been studied intensively in literature, the application towards robotics did not receive similar attention for various reasons: Images of robots present a unique set of challenges: First, robotic appearance, in contrary to human appearance, is featureless and poses difficulties in locating points of interest in pixel-space and depth. Furthermore, there are no large established datasets for robot pose estimation to be used in the context of machine learning applications.
    
\begin{figure}[!t]
    \centering
    \includegraphics[width=.9\columnwidth]{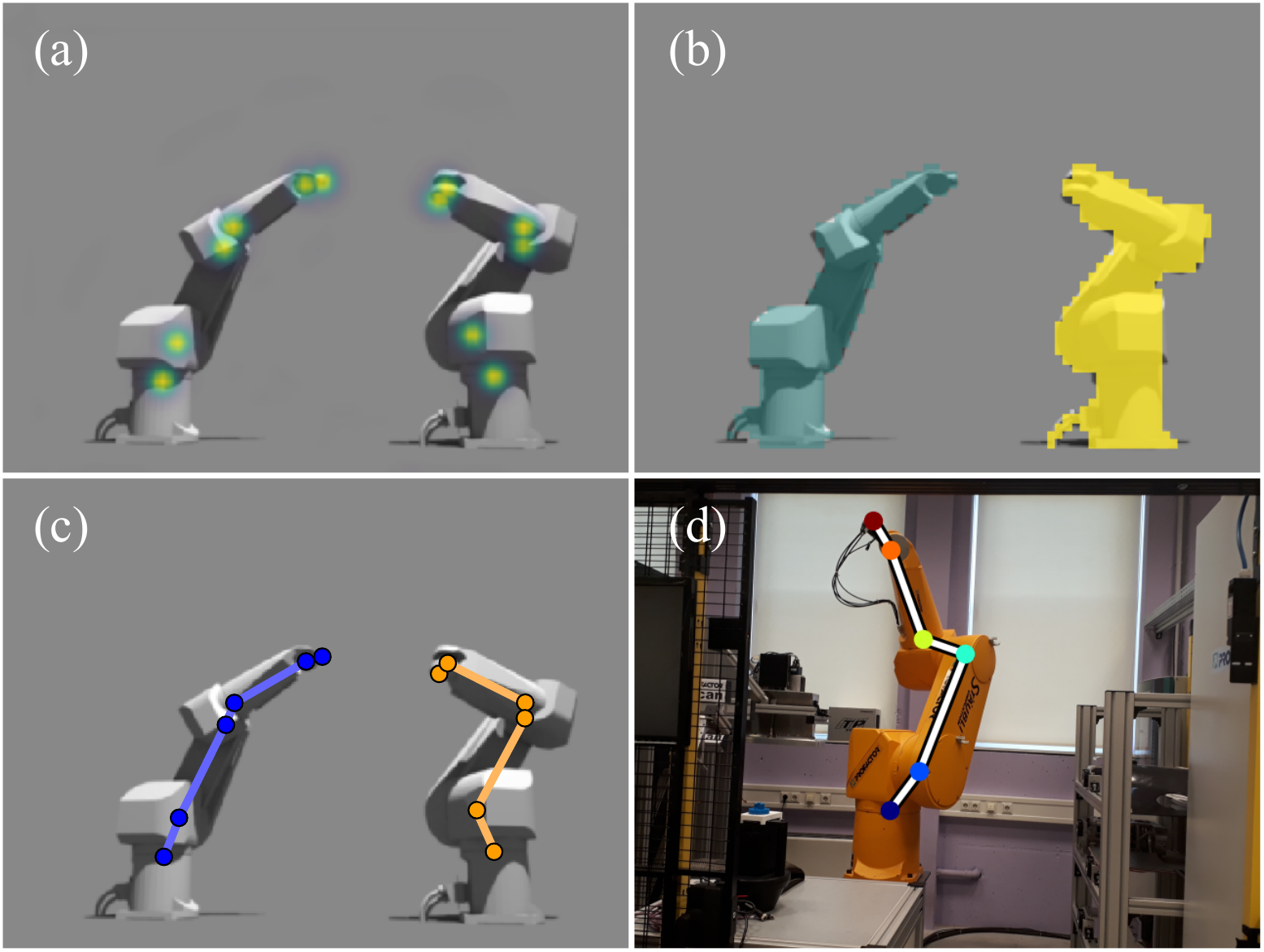}
    \caption {
        \label{fig:overview} 
        Overview: Our method predicts robot joint locations (a), robot instance segmentations (b), and - by combination of these - the complete kinematics chain per robot in 3d. We perform training on artificially rendered images and show that the method performs well on both, artificial and real data (d).
    }
\end{figure}

    In this paper we present a method for estimating articulated robot poses. Figure \ref{fig:overview} outlines our approach. Our method generates joint belief and depth maps for all robots in a single 2D input color image. Given belief and depth maps, a recurrent instance segmentation network generates robot instance masks. Finally, a kinematic chain is reconstructed by masking belief and depth maps using the instance masks, locating the most likely joint location in pixel space and re-projecting the coordinates into 3D via depth information. We harvest training data solely from a non-realistic 3D simulation. Finally, we show that our approach generalizes from simulation to real world image data. 

    The main contributions of this work are: (1) a novel approach to jointly estimate depth and location of robot joints from 2D image data, (2) an interplay with instance segmentation to avoid complex joint-to-robot inference queries, and (3) a demonstration of a pipeline containing an upstream simulation step that allows the model to generalize from artificial to real world images.
    
\section{RELATED WORK}   
    \textbf{Articulated pose estimation}, the task of identifying joint and body parts to recover pose from visual input, has been studied intensively for human kinematics in literature. Traditionally marker-less methods use a pictorial approach \cite{fischler1973representation, andriluka2009pictorial, felzenszwalb2005pictorial, pishchulin2013poselet} that represent an object by a collection of parts arranged in deformable configuration. Body part locations are predicted by individually trained detectors. The rise of Convolutional Neural Networks (CNNs) led to an increased robustness of local body part location \cite{oliveira2016deep, newell2016stacked, wei2016convolutional}. Body configurations are represented by a graph like structure with spring like forces, or probabilities attached, from which the most likely configuration is extracted using belief propagation \cite{hua2005learning, sigal2004tracking, andriluka2009pictorial, yang2013articulated}. 

   \textbf{Multi body pose detection} is typically approached in either of two ways. The top down approach to multi body pose estimation employs an object body detector followed by single pose estimation in proposal regions \cite{pishchulin2012articulated, gkioxari2014using, sun2011articulated, papandreou2017towards, toshev2014deeppose, pfister2015flowing}. Bottom up approaches locate instance independent features, such as joint locations, that are then arranged into body instances \cite{pishchulin2016deepcut, insafutdinov2016deepercut, wei2016convolutional}. Newell et al. \cite{newell2016stacked} showed that sequential prediction of body parts with intermediate loss functions helped avoiding vanishing gradients in deep network architectures. Cao et al. \cite{cao2017realtime} proposes a staged prediction network of belief and body affinity maps in which individual poses are extracted solving a bipartite graph matching problem. Papandreou et al. \cite{papandreou2018personlab} employ a bottom up approach to predict and group key-points on relative distances. The decoded poses are then used to provide an instance level segmentation of the scene.

   \textbf{Robot pose estimation} is a key ingredient in eye-to-hand visual servoing applications \cite{kanellakis2017survey} to control robots by visual input. Traditionally, these methods try to be as accurate as possible in their measurements and therefore rely on 3D depth input \cite{garcia2013guidance, varhegyivisual}. Levine et al. \cite{levine2018learning} show that hand-eye coordination can be learned in the context of grasping from monocular images. Miseikis et al. \cite{miseikis2018multi} demonstrate the usage of CNNs for predicting joint coordinates of a single robot trained on a robot specific image dataset. In a follow-up work \cite{miseikis2018transfer} they illustrated that transfer learning can significantly reduce the number of required training samples.
      
   \textbf{Deep learning from artificial images} enables machine learning on virtually unlimited amount of data. Furthermore, annotations are typically generated together with the data itself. Hence, no tedious manual labelling is required. Artificial data was recently used to perform 3D hand tracking from 2D images~\cite{mueller2018hands}. For autonomous driving 3D rendered images were used to train an image segmentation system~\cite{Johnson2017driving}. Artificial data was also used for human pose estimation~\cite{elbasiony2018deep}. Tobin et al. \cite{tobin2017domain} showed that random colorization provides a mechanism for generating artificial data that generalizes to real world objects of simple shape.
      
   In \textbf{this work} we fusion results from multi person detections and visual servoing for multi robot pose estimation based on artificial training data generated via a simulation step. Our method is similar in nature to the human body part prediction of Cao et al. \cite{cao2017realtime} and Newell et al. \cite{newell2016stacked}, but additionally estimates a sparse depth map for each joint location. It extends the method of \cite{miseikis2018transfer} to multiple robots through an instance segmentation module, that is an extension of a recurrent neural network proposed by Romera et al. \cite{romera2016recurrent}. We propose an upstream simulation step to generate artificial training data using a variant of domain randomization \cite{tobin2017domain} adapted to freeform objects. Experiments show, our method generalizes to real images of complex scenery.

\section{METHOD}
\begin{figure} [!h]
        \centering
        \includegraphics[width=.8\columnwidth]{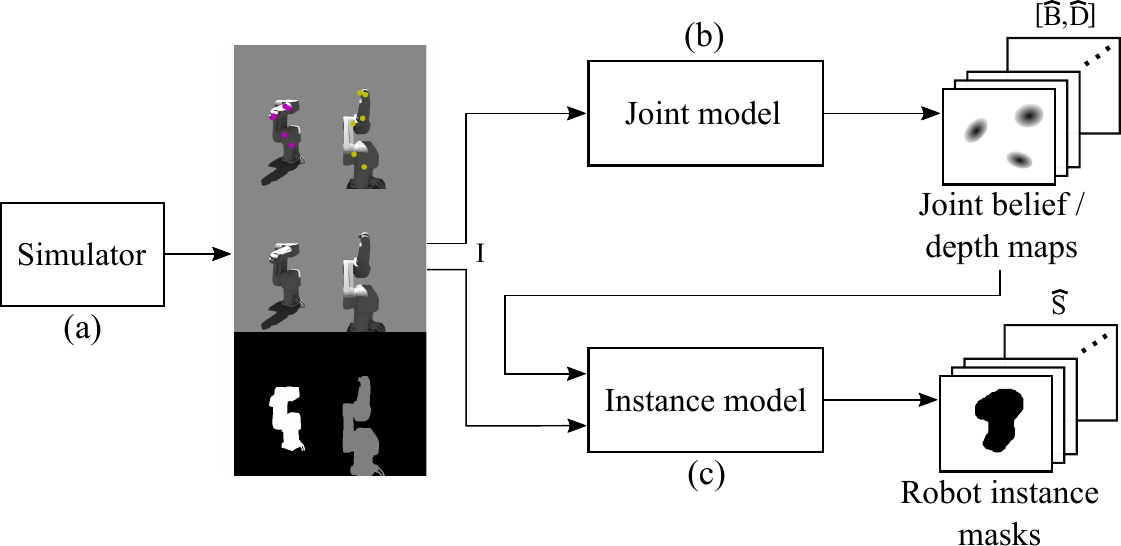}
        \caption {
            \label{fig:overviewpipeline} 
            Schematic diagram of pipeline. From a simulation the localization model predicts joint belief and depth maps. These are taken as input by the instance model, estimating robot instance masks in a recurrent fashion.
        }
    \end{figure}

    Our approach is outlined in Figure \ref{fig:overviewpipeline}. A simulator produces color images $\mathbf{I} \in \mathbb{R}^{3 \times H\times W}$ composed of a varying number of robots $R$ in random pose (Figure \ref{fig:overviewpipeline}a). Next, the joint localization model (Figure \ref{fig:overviewpipeline}b), a feed forward neural network, takes as input the style transformed images and predicts joint belief $\hat{\mathbf{B}} \in \mathbb{R}^{J \times H \times W}$, with $J$ usually being 6, and a single depth map $\hat{\mathbf{D}} \in \mathbb{R}^{1 \times H \times W}$. The results are then fed into the instance model (Figure \ref{fig:overviewpipeline}c), which estimates a set of dense pixel segmentation masks $\hat{\mathbf{S}} \in \mathbb{R}^{1 \times H \times W}$, one for per robot, using recurrent connections.

    To predict a robot pose, a segmentation mask $\hat{\mathbf{S}}$ is applied elementwise to all channels of the belief map $\hat{\mathbf{B}}$, thus narrowing down the search for a single instance. Next, individual joints are localized in each of the belief maps by applying non maximum suppression or Laplacian-of- Gaussians \cite{bretzner1998feature}. Finally, the 3D coordinates of joints are reconstructed using estimated depth $\hat{\mathbf{D}}$ and camera intrinsics $\mathbf{K}$. 

    \subsection{ARTIFICIAL DATA AND PREPROCESSING}
        Before any deep learning methods can be applied, a sufficiently large amount of training data needs to be available. For the domain at hand it is impractical to rely on manually annotated real data and to the best of our knowledge no such datasets exist. 
        
        \begin{figure} [!h]
            \centering
            \includegraphics[width=.9\columnwidth]{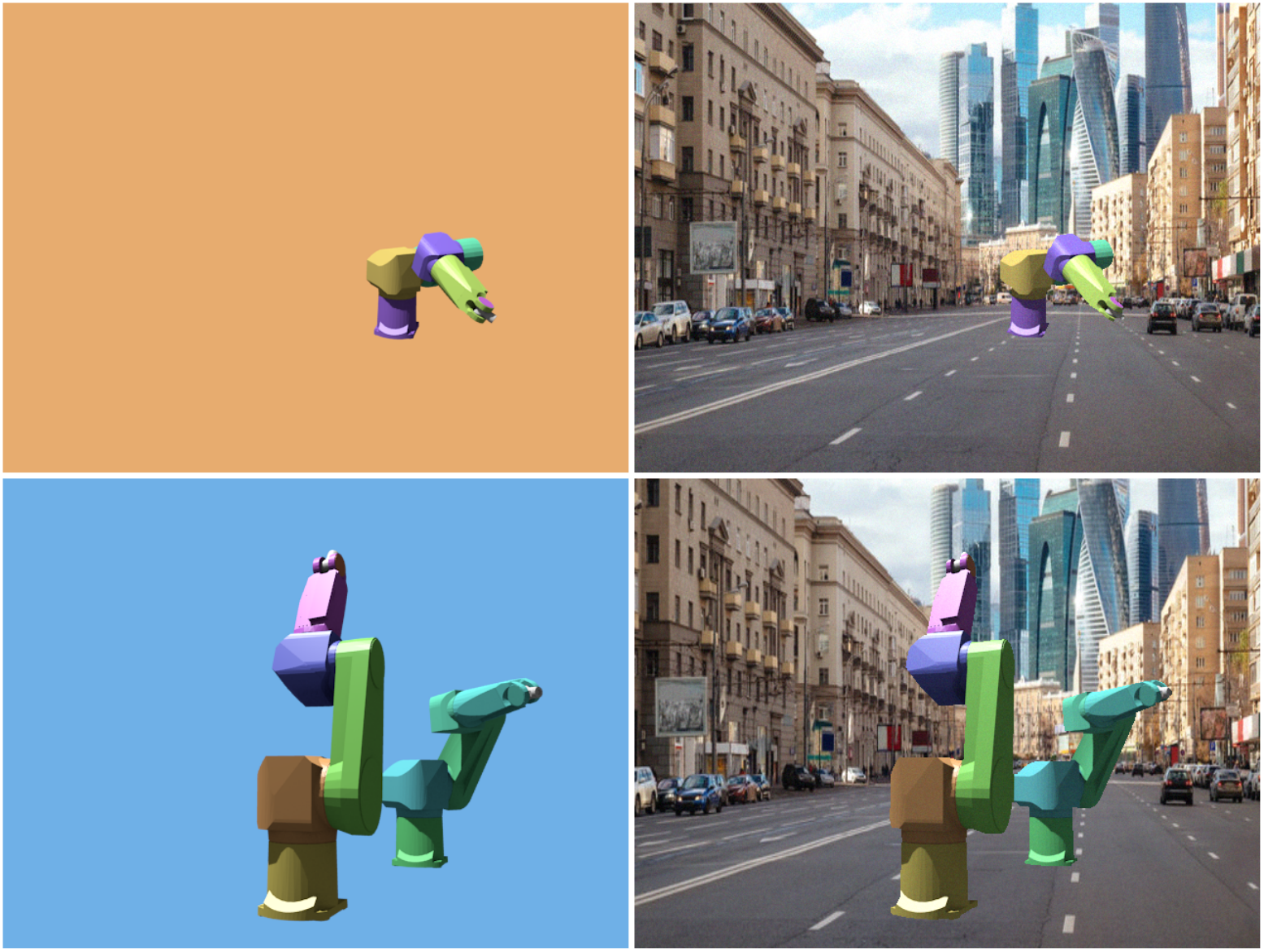}
            \caption {
                \label{fig:artificialImages} 
                Two examples of artificially generated images. Each row shows a separate example (top row with two robots visible, bottom row with single robot visible). The left column shows images generated by the rendering engine. For illustration, each joint is randomly colored in this example. The right column shows training input images with real photo in the background. The same background image is used in both examples (differently scaled and cropped).
            }
        \end{figure}
        
        We therefore propose to use artificially generated images of robots. We make use of the freely available Blender software to generate a large number of training images. Figure \ref{fig:artificialImages} illustrates how artificial images are generated. The robots are first rendered in {3D} with random configurations. Then, real world images are added as background. The 3D rendering pipeline produces rendered RGB color images $\mathbf{I}$, segmentation masks $\mathbf{S}$, and 2D pixel joint locations (used to generate ground truth belief maps $\mathbf{B}$) as well as 3D joint locations (used to generated ground truth depth maps $\mathbf{D}$).
        
        There is a set of parameters that are randomly chosen to generate a large variety of training images:
        \begin{itemize}
            \item{Robot visibility: A random variable $v_r \sim \mathrm{Bernoulli}(\theta_\mathrm{vr})$ with hyper-parameter $\theta_\mathrm{vr}$ controls the global visibility of each robot and is sampled per robot per frame.}
            \item{Robot color: The diffuse color of every robot, or every joint, is sampled uniformly $\mathrm{Uniform}(0,n_{\mathrm{color}})$ from a set of $n_{\mathrm{color}}$ linear spaced colors in HSV space.}
            \item{Camera position: The virtual camera with intrinsic parameters $\mathbf{K} \in \mathbb{R}^{3\times 3}$ is randomly positioned on a hemisphere with radius $c_r \sim \mathrm{Uniform(0,\mathrm{r_{max}})}$, inclination angle $c_\theta \sim \mathrm{Uniform(0,\pi)}$ and azimuth angle $c_\phi \sim \mathrm{Uniform(0,2\pi)}$. Each camera is directed towards the center of the scene.}
            \item{Joint angles: Poses of the robot are randomly chosen by assigning random joint angles $\alpha_j$ to each of the joints $j$ with $\alpha_j \sim \mathrm{Uniform(-\pi/2, \pi/2)}$}.
            \item{Background image: {3D} renderings of the robots in different poses are masked and drawn on top of real photos. Background images are randomly chosen from a fixed pool of images. Each background is randomly scaled and cropped.}
            \item{Image noise: Some noise is added to the final generated images. The amplitude $a$ is randomly chosen with uniform distribution $a \sim \mathrm{Uniform(0, 30)}$ (with pixel values ranging from 0 to 255).}
        \end{itemize}

        Besides the artificial images, which serve as input to the neural networks, it is possible to extract other useful data from the rendering pipeline. Essentially, we store 3d robot joint locations and masks of rendered robots in the images. This data is fed as ground truth into the loss functions that are used for neural network training. 

    \subsection{LOCALIZATION}
\begin{figure} [!h]
            \centering
            \includegraphics[width=0.98\columnwidth]{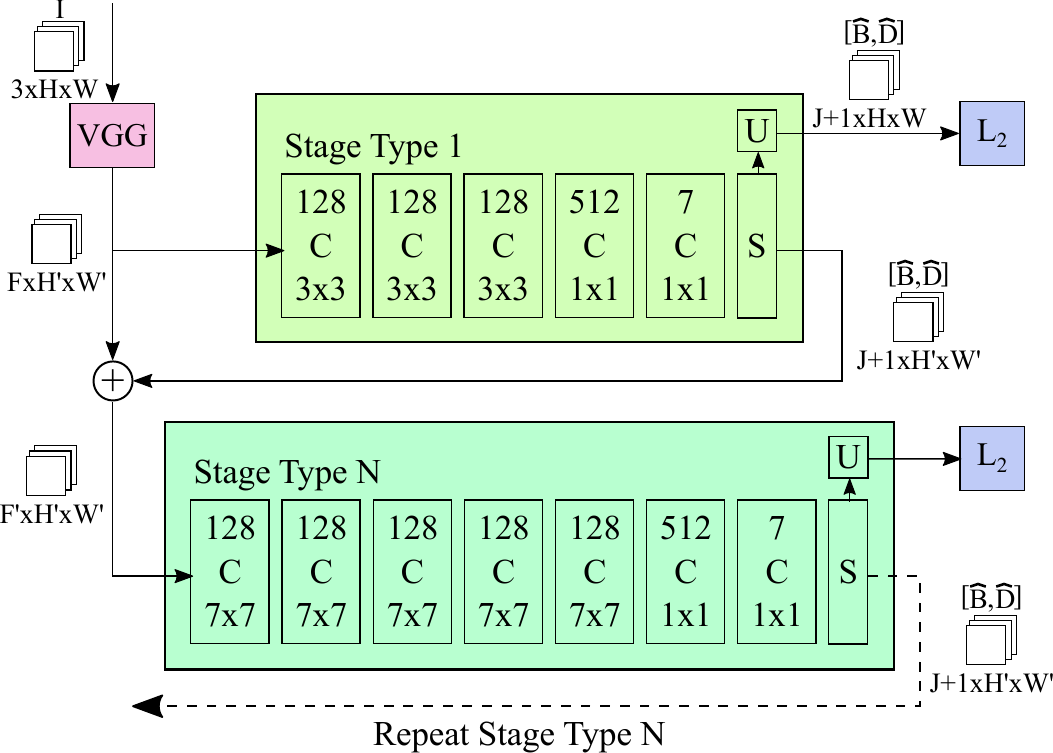}
            \caption {
                \label{fig:jointmodel} 
                Architecture of the localization module. Each stage predicts joint belief maps and depth values in close vicinity to joint locations. Successive stages are fed a channel concatenated stack of VGG features and the current estimate. Here C is block consisting of (Conv, BN, ReLU) with number of output features written above and kernel size below, S is a sigmoid layer, and U a block composition of nearest neighbor interpolation, C blocks (3x3) followed by S.
            }
        \end{figure}

        The details of the localization model are shown in Figure \ref{fig:jointmodel}. Its task is the prediction of robot joint beliefs $\hat{\mathbf{B}}$ and depth maps $\hat{\mathbf{D}}$ from color image input $\mathbf{I}$. Our approach, in style of Wei et al. \cite{wei2016convolutional}, iteratively refines its estimates via a successive application of fully convolutional stages \footnote{Details of our architecture can be found at \url{https://github.com/cheind/robot-pose}}.

        First, color images are fed through a pre-trained VGG \cite{simonyan2014very} network to generate base features of size $\mathbb{R}^{F \times H' \times W'}$. Next, each stage simultaneously predicts joint belief and single depth maps $\mathbb{R}^{J+1 \times H' \times W'}$. Both, belief and depth image entries are in the $\left[0,1\right]$ range, naturally produced by a sigmoid output layer at each stage. Additionally, an up-sample filter using nearest-neighbor interpolation followed by convolutional elements, transforms outputs to color image dimensions $\mathbb{R}^{J+1 \times H \times W}$. Except for the first stage, inputs are composed of a channel concatenated version of outputs of the previous stage and base features. 

        During training, we apply two losses after each stage to the up-sampled outputs. The first loss $L_2^B$ directly penalizes the squared error between predicted $\hat{\mathbf{B}}$ and target joint belief maps $\mathbf{B}$. Regarding the depth loss, target depth values $\mathbf{Z}$ are transformed to the $\left[0,1\right]$ range by applying an inverse depth $\mathbf{D}=a\frac{1}{\mathbf{Z}} + b$ transformation \cite{hartley2003multiple}.

        The complete training loss for stage $n$ in the joint localization model is given by
        \begin{equation}
            L_n = L_2^B + \sum\limits_{\mathrm{p} \in \Omega}\mathbf{W}_{\mathrm{p}}\left(\mathbf{D}_{\mathrm{p}} - \hat{\mathbf{D}}_{\mathrm{p}}\right)^2
        \end{equation}
        where $\Omega \in \mathbb{R}^{H\times W}$ is the image domain and $\mathbf{W}_{\mathrm{p}}$ is a pixel wise weight given as a function of the maximum belief at pixel $\mathrm{p}$
        \begin{equation}
            \mathbf{W}_{\mathrm{p}} = \left\{
                \begin{array}{ll}
                1 & \textrm{if } \max\limits_{j\in{J}}{\mathbf{B}}_{\mathrm{j,p}} \ge \phi \\
                0 & \textrm{else}
                \end{array}
            \right\}.
        \end{equation}
        Here $\phi$ is a tunable hyper-parameter that controls the radius of a disc of constant depth centered at joints. The total loss for the joint model accumulates over all stages
        \begin{equation}
            L = \sum\limits_{i=1}^{N}L_i.
        \end{equation}
        When predicting from the model, only the result from the last stage is used. in training the intermediate loss signals at every stage help addressing the vanishing gradient problem \cite{wei2016convolutional}.
        
    \subsection{INSTANCE SEGMENTATION}
\begin{figure} [!h]
            \centering
            \includegraphics[width=\columnwidth]{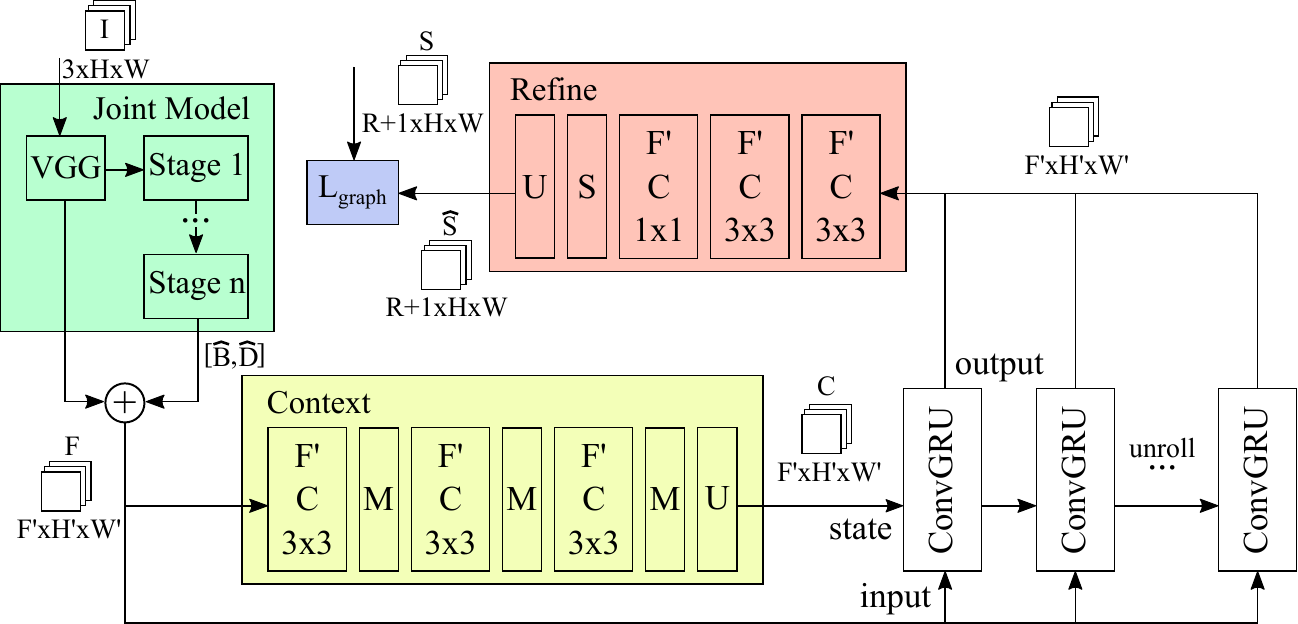}
            \caption {
                \label{fig:instancemodel} 
                Architecture of the instance segmentation module unrolled for $R+1$ iterations. Given VGG features and joint prediction concatenated into the tensor $\mathbf{F}$, the system generates pixel-wise segmentation masks by refining the output of convolutional gated recurrent units (ConvGRU). A context module down-samples $\mathbf{F}$ to provide global information as initial state seed. The cell also takes $\mathbf{F}$ as input in every iteration. Abbreviations explained in Figure \ref{fig:jointmodel}, except for M which refers to max-pooling.
            }
        \end{figure}

        The instance model, shown in Figure \ref{fig:instancemodel}, follows a recurrent architecture using convolutional gated recurrent units (ConvGRU) \cite{ballas2015delving}. Convolutional recurrent cells replace fully connected layers of conventional GRUs by convolutional operators. This effectively makes them independent of lateral input dimensions, exploits the nature of convolutional inputs and reduces the number of parameters. 

        The model input $\mathbf{F} \in \mathbb{R}^{J+1+F \times H' \times W'}$ is composed of localization predictions $\hat{\mathbf{B}}$ and $\hat{\mathbf{D}}$ concatenated with VGG base features of the input image $\mathbf{I}$. The context module down-samples $\mathbf{F}$ via a sequence of convolution blocks followed by max pooling operations. It provides global image context, in spirit of the contraction paths in U-Nets \cite{ronneberger2015u}. Its output $\mathbf{C} \in \mathbb{R}^{J+1+F \times H' \times W'}$ forms the initial ConvGRU state seed. The input to the ConvGRU cell is $\mathbf{F}$ in all iterations. The cell's output is a low-resolution feature map that is then fed into a fully convolutional refine block generating the final instance segmentation mask $\hat{\mathbf{S}} \in \mathbb{R}^{1 \times H \times W}$.

        During training the number of robots $R$ and target masks $\mathbf{S}$ are provided. We augment $\mathbf{S}$ by an end-of-sequence mask without any pixel activation. Thus, let $\mathbf{S}' \in \mathbb{R}^{R+1 \times H \times W}$. Given $\mathbf{I}$ and access to the localization network, we predict $\hat{\mathbf{S}} \in \mathbb{R}^{R+1 \times H \times W}$. The instance segmentation loss $L_\textrm{graph}$ is composed of a term penalizing robot instance segmentation errors and end-of-sequence mismatches. To account for the permutation invariance of robot instances we solve a minimum weight matching problem in bipartite graphs \cite{kuhn1955hungarian} between the first $R$ masks of $\mathbf{S}'$ and $\hat{\mathbf{S}}$. The cost associated for assigning instance mask $\mathbf{S}'_i$ to $\hat{\mathbf{S}}_j$ is computed by a continuous version of the intersection-over-union metric \cite{krahenbuhl2013parameter} given by
        \begin{align}
            k &= \langle\mathbf{S}'_i,\hat{\mathbf{S}}_j\rangle \\
            C_{ij} &= 1 - \frac{k}{\sum\limits_{\mathrm{p} \in \Omega}\mathbf{S}'_{i,p} + \sum\limits_{\mathrm{p} \in \Omega}\hat{\mathbf{S}}_{j,p} - k}
        \end{align}
        where the inner product $\langle\cdot,\cdot\rangle$ is applied to instance masks reshaped as column vectors. For the end-of-sequence mask at index $R+1$ the standard squared error loss $L_2$ is applied to target and prediction masks. The combined loss for instance segmentations is thus
        \begin{equation}
            L_{\mathrm{graph}} = \min\limits_{\mathbf{A}}\sum\limits_{i=1}^R\sum\limits_{j=1}^R C_{ij}A_{ij} + L_2\left(\hat{\mathbf{S}}_{R+1}, \mathbf{S}'_{R+1}\right)
        \end{equation}
        where $\mathbf{A} \in \{0,1\}^{R \times R}$ is Boolean matrix with $\mathbf{A}_{ij}=1$ if target robot $i$ is assigned to prediction $j$. Constrained to each row must assigned to at most one column and vice versa. 

\section{RESULTS}

    In the following we describe the training procedure, evaluation of our method on synthetic and real world data, and provide runtime metrics.
    
    \subsection{TRAINING}
        For training of joint and instance models, a set of 10.000 random robot scenes were generated with the following parameters: $\theta_\mathrm{vr}=0.75$, $n_{\mathrm{color}}=400$, $r_{max}=\SI{10}{\metre}$. Two distinct collections of random background images (city and industrial theme) were used as synthetic training- and test-set backgrounds. We trained the joint localization model using five stages with input images down-sampled to $480 \times 640$. A VGG network, pre-trained on detection, was used to generate base features. We used Adam \cite{kingma2014adam}, $\eta=\num{1e-3}$, optimization with mini-batch learning for a total of 30 epochs.
    
    \subsection{EVALUATION}
\begin{figure}[!h]
            \centering
            \begin{subfigure}[t]{0.49\columnwidth}
                \includegraphics[width=\columnwidth]{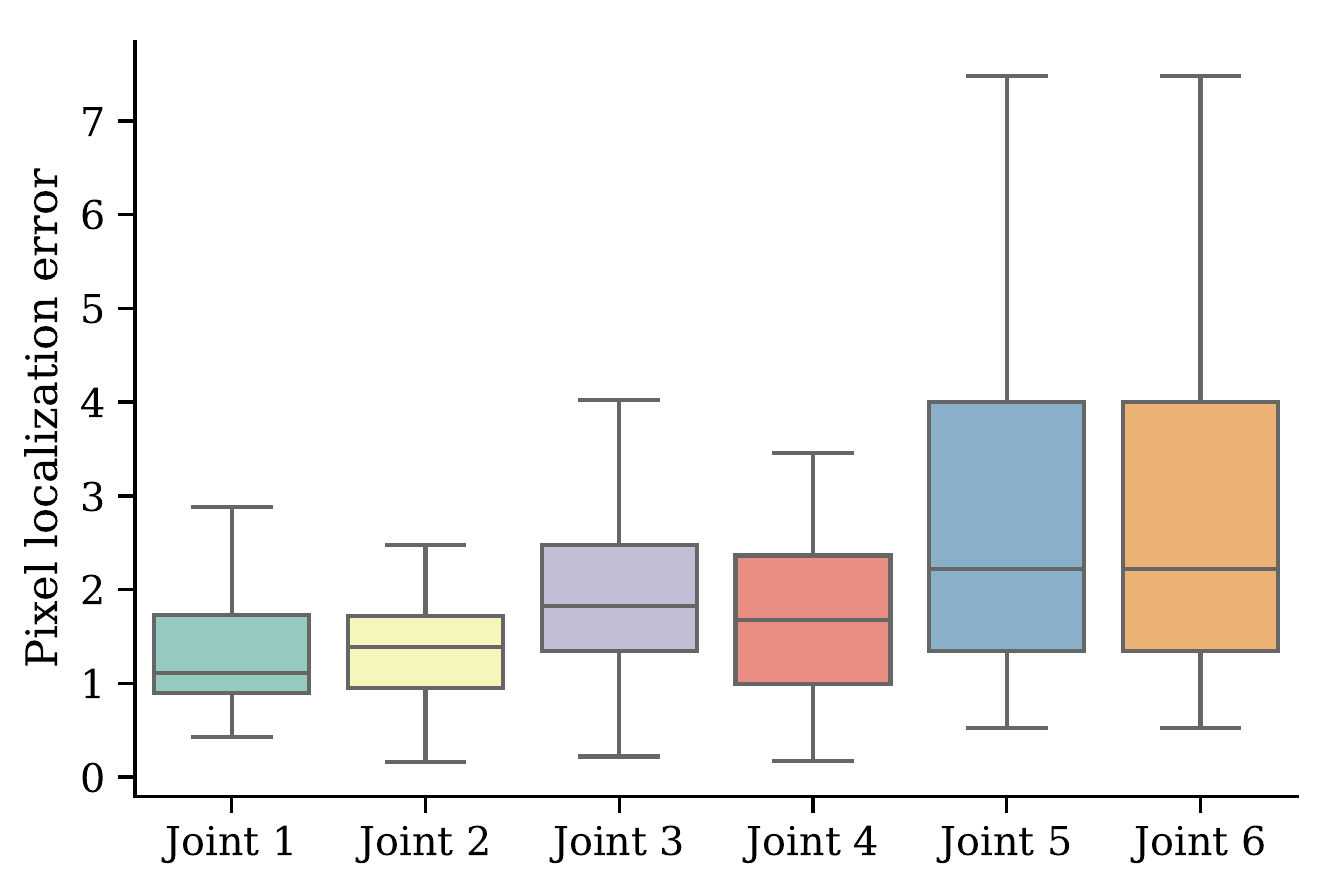}
                \caption {
                    \label{fig:pixelerrornms} 
                    Pixel errors of joint prediction (non maxima suppression).
                }
            \end{subfigure}
            \begin{subfigure}[t]{0.49\columnwidth}
                \includegraphics[width=\columnwidth]{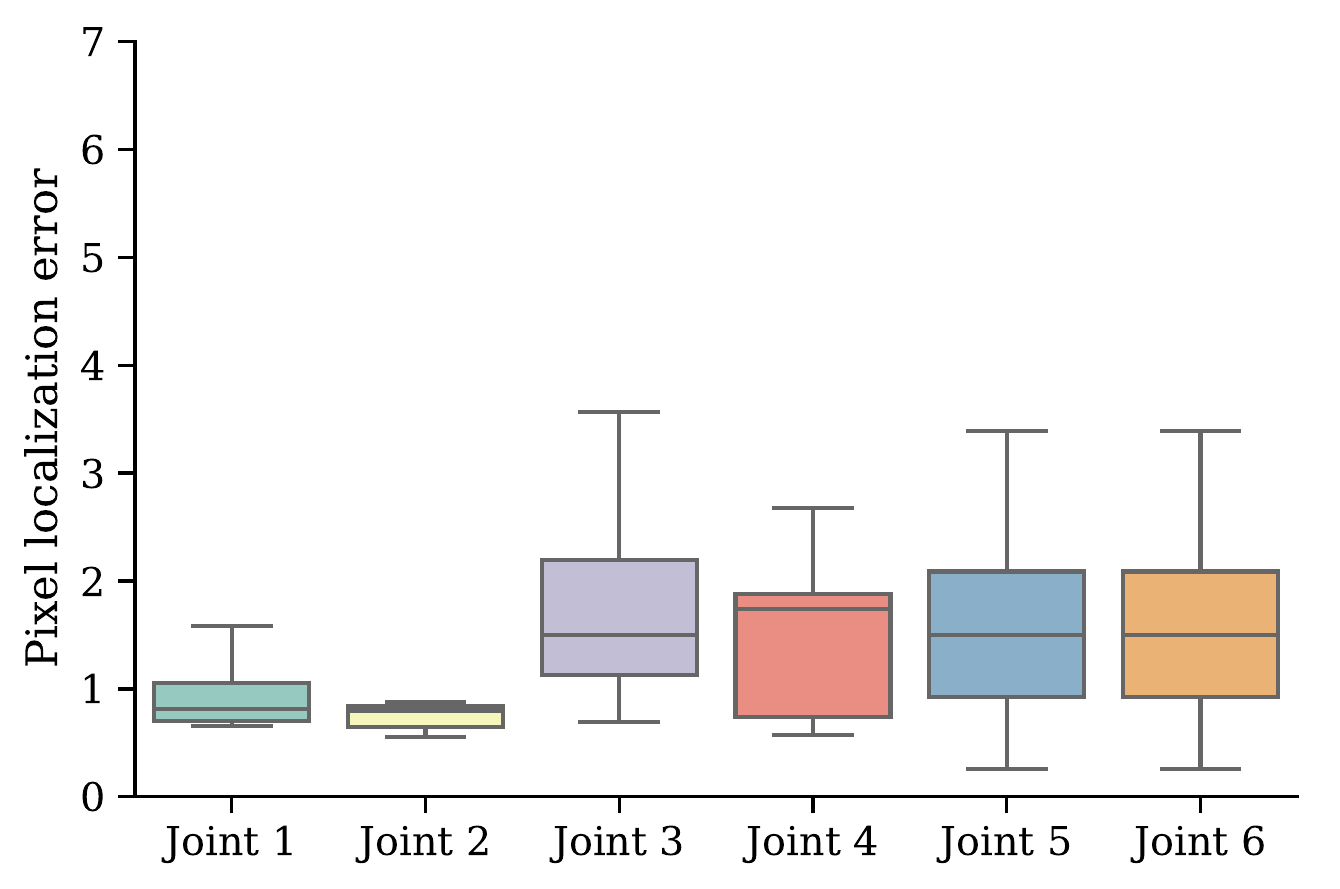}
                \caption {
                    \label{fig:pixelerrorlog} 
                    Pixel errors of joint prediction (Laplacian of Gaussian).
                }
            \end{subfigure}
            \caption {
                \label{fig:pixelerror} 
                Pixel errors of joint prediction from belief maps on artificial training data.
            }
        \end{figure}

        We first evaluate the joint model on the artificial test data set with respect to pixel and depth errors. We define the pixel localization error as the Euclidean distance between predicted and target coordinates. Figure \ref{fig:pixelerror} compares the errors of two joint prediction methods. The estimation of the base joints (1-2) is considerably more accurate then the prediction of the later joints. Especially the fifth and sixth joint, located closely at the robot tip, are harder to detect precisely. We reason that this is because later joints show more movement, orientation and are smaller in image space compared to base joints. Our findings are also observed in Miseikis et al. \cite{miseikis2018multi}. 

\begin{figure}[!h]
            \centering
            \begin{subfigure}[t]{0.49\columnwidth}
                \includegraphics[width=\columnwidth]{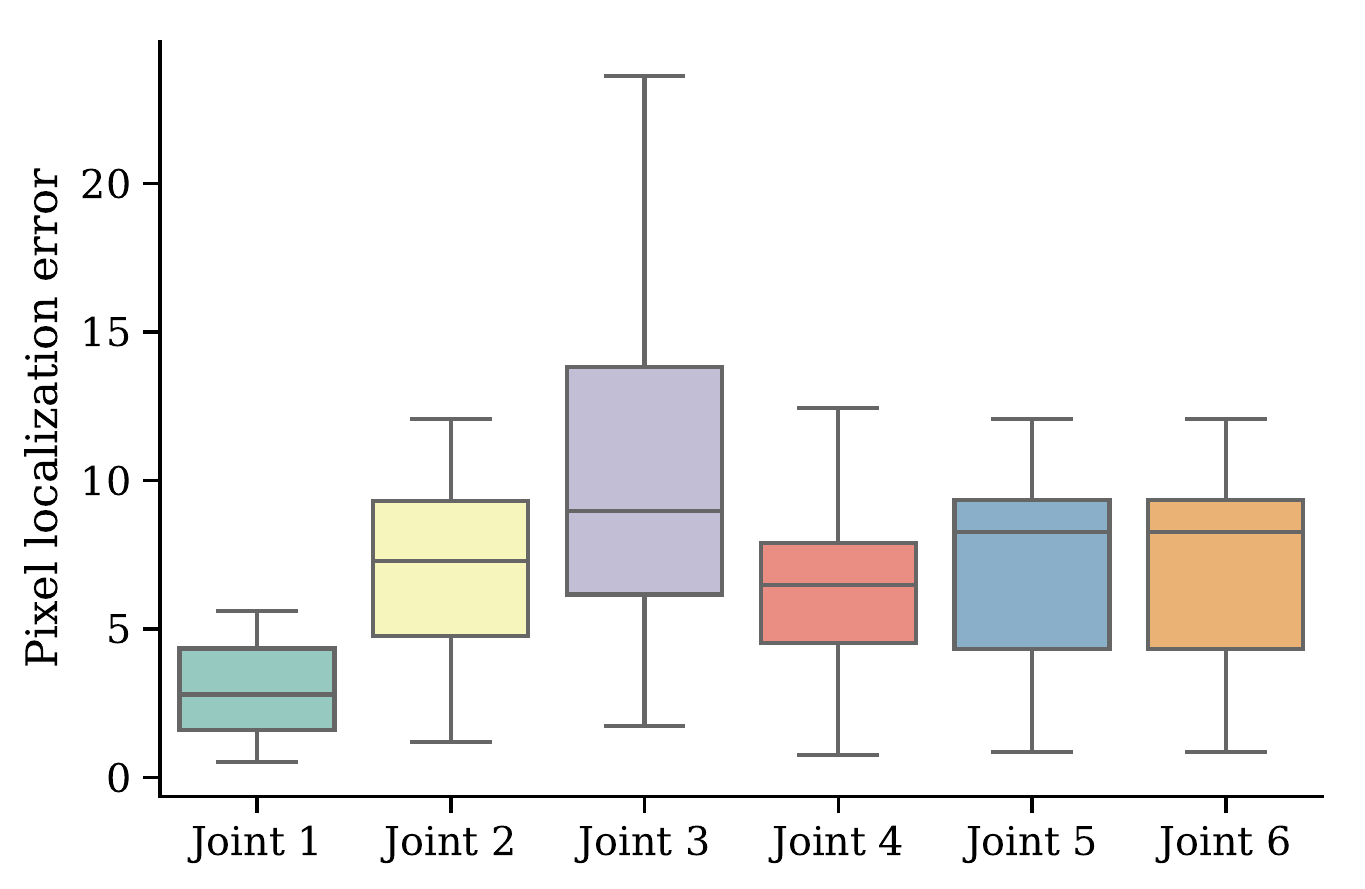}
                \caption {
                    \label{fig:pixelerrorreal} 
                    Pixel errors of joint prediction for real world images.
                }
            \end{subfigure}
            \begin{subfigure}[t]{0.49\columnwidth}
                \includegraphics[width=\columnwidth]{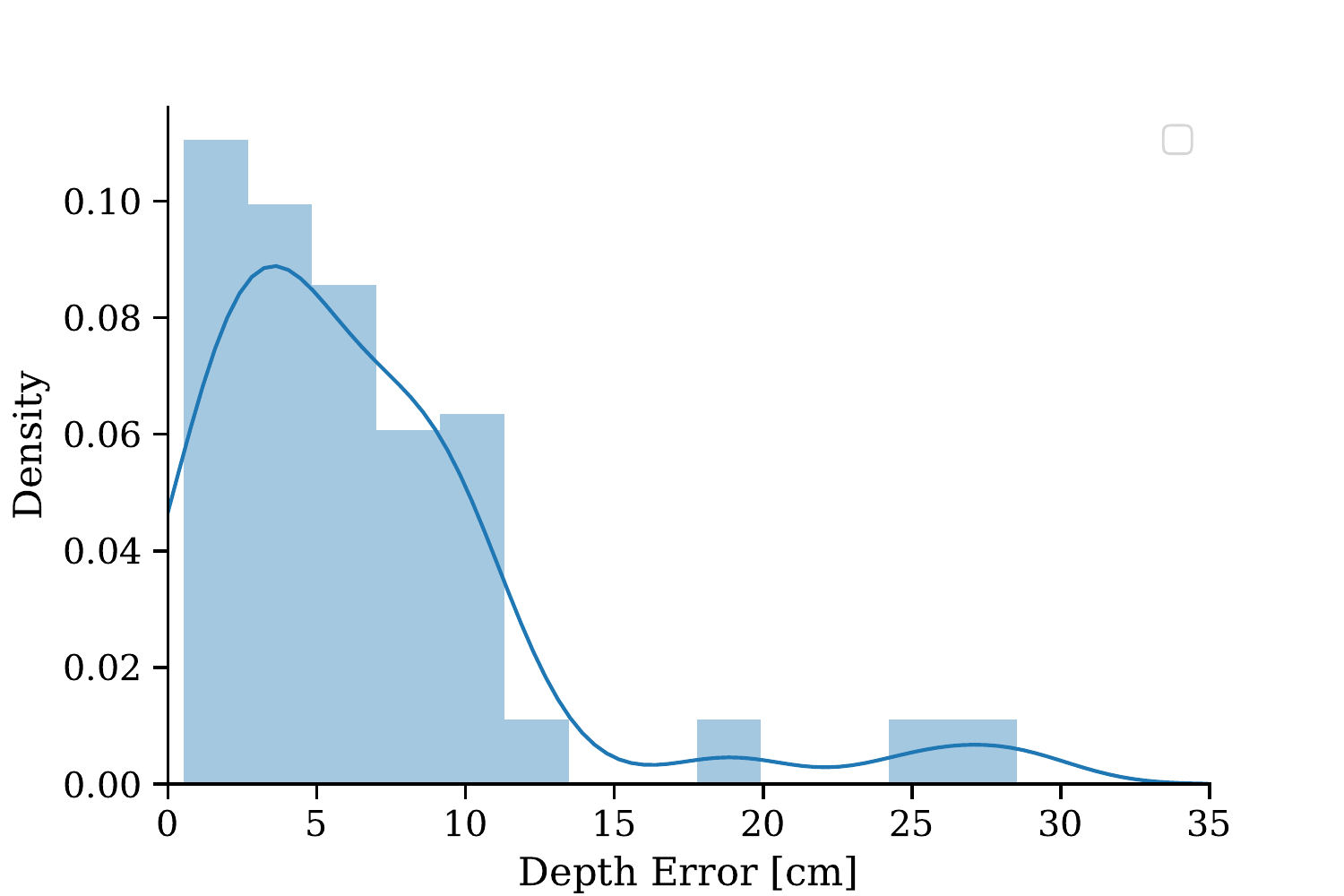}
                \caption {
                    \label{fig:deptherrorhist} 
                    Histogram of depth errors in $\SI{}{\cm}$ of artificial data with kernel density estimation.
                }
            \end{subfigure}
            \caption{
                Joint localization and depth errors.
            }
        \end{figure}

        Depth accuracy is only calculated for regions where ground truth belief maps are above threshold $\phi$. We subdivide the maximum depth of \SI{10}{\metre} into a near and far range and evaluate separately. Table \ref{tab:deptherror} summarizes the depth errors for the artificial test set in centimeters averaged over all joints. Details are shown in figure \ref{fig:deptherrorhist}. The results are comparable to the findings in Miseikis et al. \cite{miseikis2018multi}, however we note that our distance in training and evaluation is considerably larger (by factor of 2). Unfortunately their dataset was not accessible to us at the time of preparing the paper for comparative results.

        \begin{table}[!h]
            \centering
            \begin{tabular}{lrr}
                \toprule
                Range &       near [$\SI{}{\cm}$] &  far [$\SI{}{\cm}$]  \\
                \midrule
                Mean  &    5.81 &   9.50 \\
                Std   &    5.07 &   7.74 \\
                25\%   &    2.52 &   4.05 \\
                50\%   &    5.12 &   6.09 \\
                75\%   &    8.35 &  12.34 \\
                \bottomrule
            \end{tabular}
            \caption{
                \label{tab:deptherror} 
                Errors of depth estimates. The maximum depth of \SI{10}{\metre} is divided into two ranges, near $[0,5)$\SI{}{\metre} and far $[5,10]$\SI{}{\metre}.
            }
        \end{table}

\begin{figure}[!h]
            \centering
            \begin{subfigure}[b]{0.64\columnwidth}
                \includegraphics[width=\columnwidth]{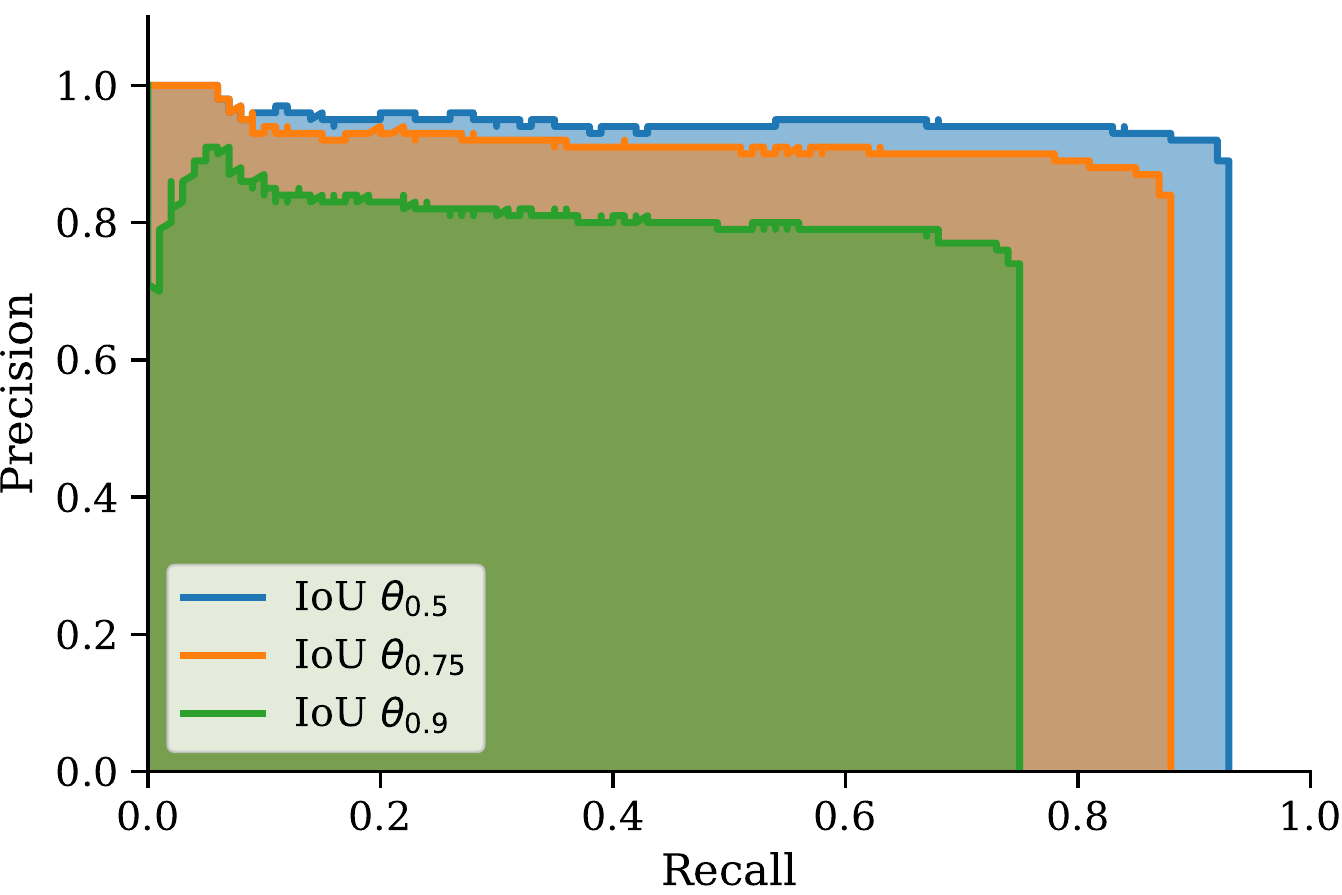}
                \caption {
                    \label{fig:precisionrecallcurve} 
                    Precision-recall curves.                    
                }
            \end{subfigure}
            \begin{subfigure}[b]{0.34\columnwidth}
                \begin{tabular}{rr}
                    \toprule
                    IoU &    AP \\
                    \midrule
                    $\theta_{0.5}$ &  88.7\% \\
                    $\theta_{0.75}$ &  81.1\% \\
                    $\theta_{0.9}$ &  61.7\% \\
                    \bottomrule
                    \end{tabular}
                \caption{
                    \label{tab:mAP} 
                    Mean precision.
                }
            \end{subfigure}
            \caption{
                \label{fig:precisionrecall} 
                Average precision at varying levels of intersection-over-union (IoU) thresholds.
            }
        \end{figure}

        We evaluate the instance model on artificial test data using the average precision (AP) metric \cite{everingham2010pascal}, measuring the maximum precisions at different recall values. For each scene we iterate the model as long as at least a fraction (5\%) of pixels are activated. We report the average pixel activation as a measure of confidence for our predictions. Since the AP requires rectangular boxes for evaluation, we generate one bounding rectangle per instance from pixels with activation level above $0.6$. Figure \ref{fig:precisionrecallcurve} shows precision recall curves for varying thresholds of intersection-over-union (IoU). Table \ref{tab:mAP} summarizes the average precision.

\begin{figure} [!h]
    \centering
    \includegraphics[width=0.98\columnwidth]{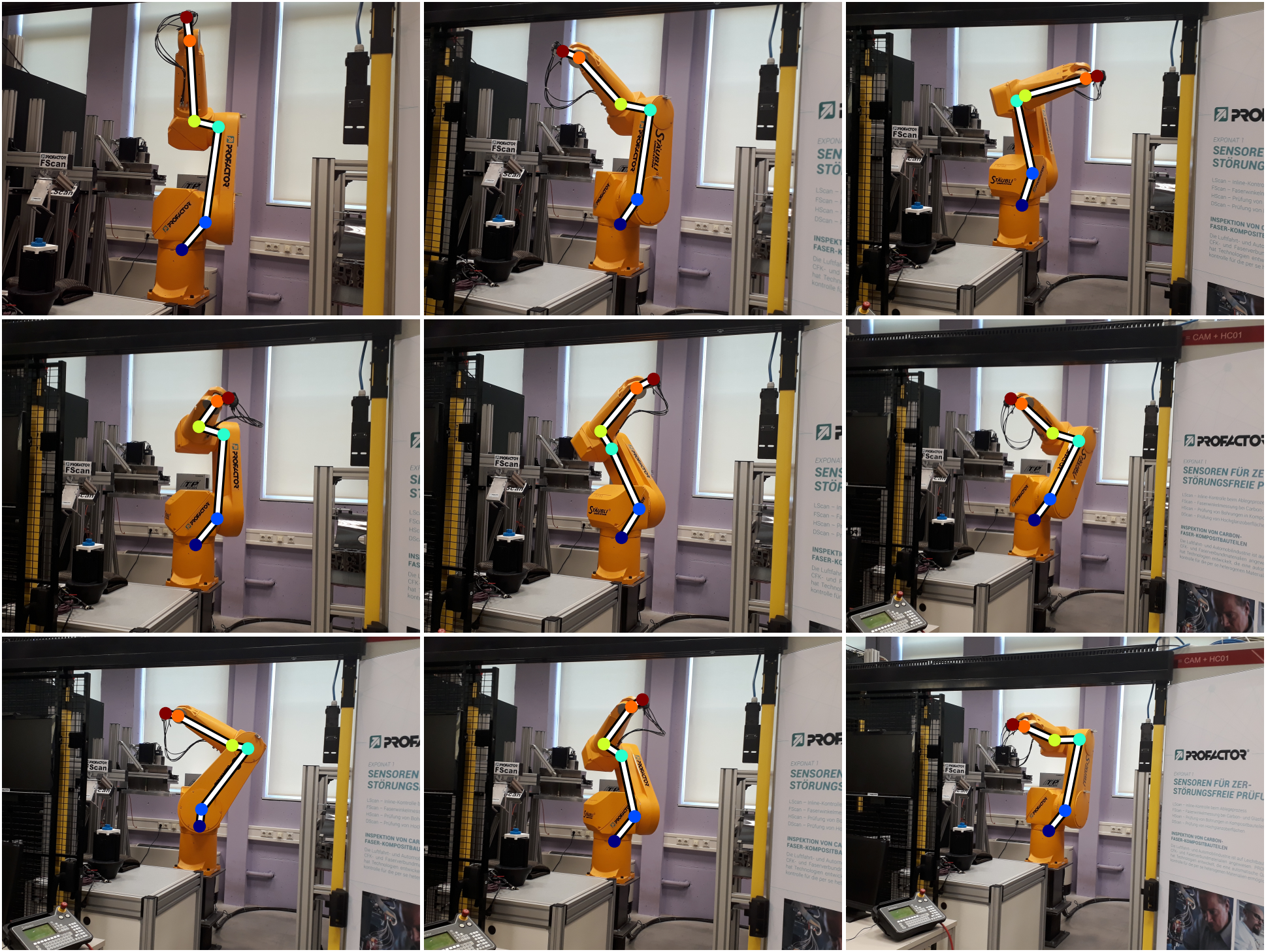}
    \caption {
        \label{fig:resultsreal} 
        Examples for robot joint localization on real test data: Located robot joints in 2d images.
    }
\end{figure}

        Besides artificial data, we evaluated our approach on real world images of robots. We collected a set of 23 images of a Stäubli RX130 robot from our lab in random poses. We annotated joint coordinates by hand and computed pixel errors as before (see Figure \ref{fig:pixelerrorreal}). The results indicate that our method generalizes to real world data, despite training was performed on outputs of a simple, unrealistic simulation step. In Figure \ref{fig:resultsreal} we superimpose the detected kinematic chain on top of input images.
        
    \subsection{RUNTIME CONSIDERATIONS}
        We run all experiments on an desktop PC with Nvidia GTX 1080i GPU, INTEL i7-8700 CPU, and 16GB RAM. Table \ref{tab:runtime} summarizes the individual runtimes of different parts of our pipeline. Belief prediction is independent of the number of robot instances and real-time capable. Retrieving a single segmentation mask is fast as well, but bound by the number of instances in the image. Extracting joint coordinates through non maxima suppression can leverage GPU power through max-pooling operations. For Laplacian of Gaussian we resorted to an untuned CPU implementation.

        \begin{table}
            \centering
            \begin{tabular}{lrr}
                \toprule
                {} &  Mean [ms] &  Std [ms] \\
                Task                   &            &           \\
                \midrule
                Belief prediction      &      10.95 &    12.675 \\
                Instance prediction    &      17.75 &     0.625 \\
                Joint localization NMS &       0.41 &     0.125 \\
                Joint localization LoG &     310.00 &     3.225 \\
                \bottomrule
            \end{tabular}
            \caption{
                \label{tab:runtime} 
                Runtimes of different parts of our pipeline for $480 \times 640$ input images. Timings in milliseconds.
            }
        \end{table}



\begin{figure} [!h]
    \centering
    \includegraphics[width=0.98\columnwidth]{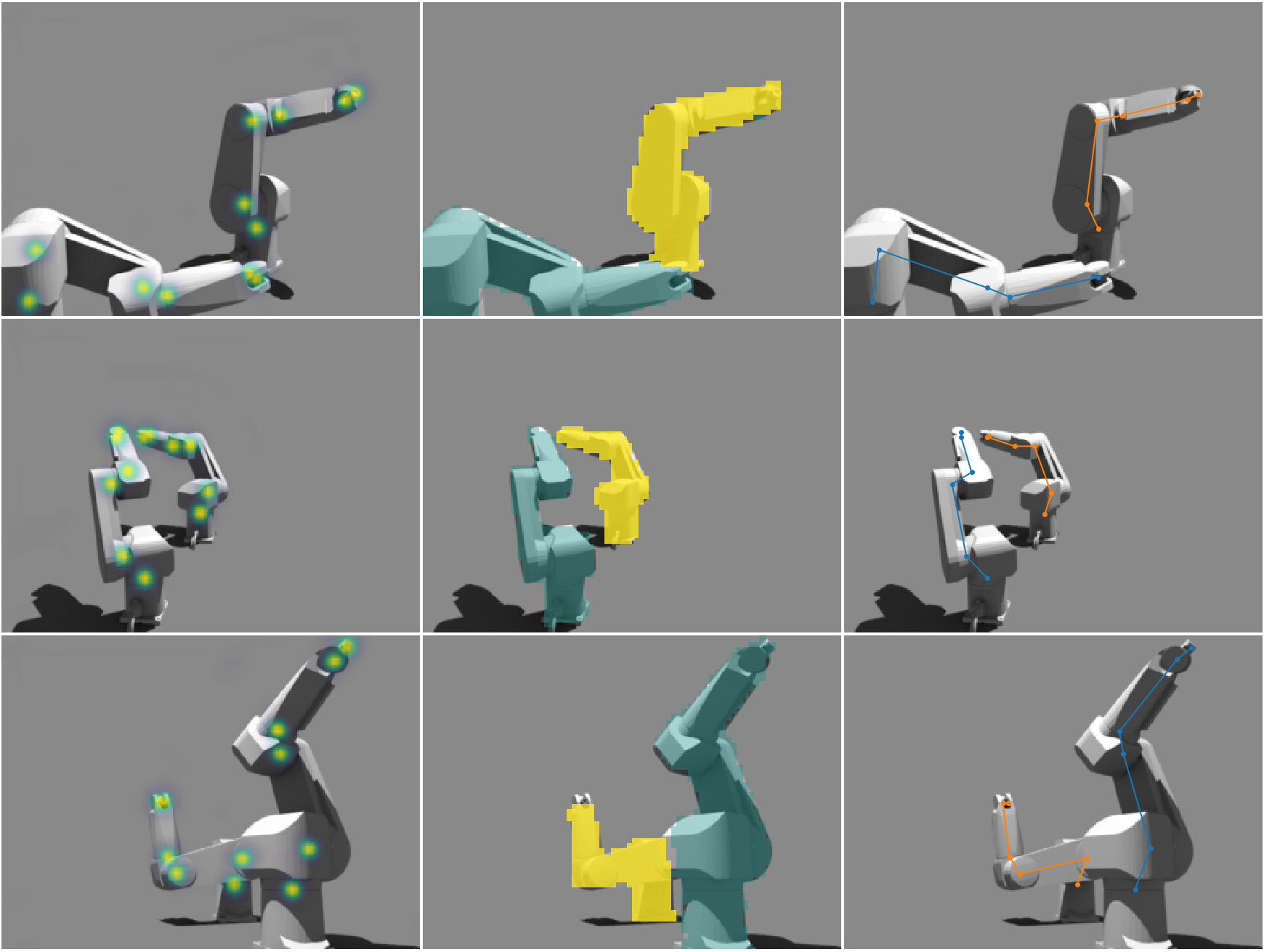}
    \caption {
        \label{fig:resultsartificial} 
        Examples for detected joints on artificial test data: Heat maps of detected joint locations (left), generated masks from instance segmentation (center), and robot kinematic chains after combination of joint heat maps and instance segmentations (right).
    }
\end{figure}

\section{DISCUSSION}
    In this work we consider a novel approach for simultaneously localizing and segmenting robot instances in 2D images. We show that harvesting training data from a non-realistic simulation suffices to generalize from artificial training data to real world images. Thereby we completely avoid a costly data collection and annotation step. 
    
    \textbf{Criticsm}
    Considering the instance model, we find that it is rather tricky to train and its runtime is lower bounded by the number of instances in the image. In future work, we will address alternative methods for instance segmentation to avoid these issues. 
    
    Further, we plan to extend our experiments to different types of robots in the wild. Such experiments require a standardized robot pose estimation dataset for traceable and reproducible results, which is unfortunately currently unavailable to the research community.

\section*{ACKNOWLEDGMENT}
    This research was supported in part by Lern4MRK (Austrian Ministry for Transport, Innovation and Technology), ”FTI Struktur Land Oberoesterreich (2017-2020)”, the European Union in cooperation with the State of Upper Austria within the project Investition in Wachstum und Beschäftigung (IWB).

{
\small
\bibliographystyle{ieeetr}
\bibliography{biblio}

\begin{thebibliography}{10}

\bibitem{fischler1973representation}
M.~A. Fischler and R.~A. Elschlager, ``The representation and matching of
  pictorial structures,'' {\em IEEE Transactions on computers}, vol.~100,
  no.~1, pp.~67--92, 1973.

\bibitem{andriluka2009pictorial}
M.~Andriluka, S.~Roth, and B.~Schiele, ``Pictorial structures revisited: People
  detection and articulated pose estimation,'' in {\em Computer Vision and
  Pattern Recognition, 2009. CVPR 2009. IEEE Conference on}, pp.~1014--1021,
  IEEE, 2009.

\bibitem{felzenszwalb2005pictorial}
P.~F. Felzenszwalb and D.~P. Huttenlocher, ``Pictorial structures for object
  recognition,'' {\em International journal of computer vision}, vol.~61,
  no.~1, pp.~55--79, 2005.

\bibitem{pishchulin2013poselet}
L.~Pishchulin, M.~Andriluka, P.~Gehler, and B.~Schiele, ``Poselet conditioned
  pictorial structures,'' in {\em Computer Vision and Pattern Recognition
  (CVPR), 2013 IEEE Conference on}, pp.~588--595, IEEE, 2013.

\bibitem{oliveira2016deep}
G.~L. Oliveira, A.~Valada, C.~Bollen, W.~Burgard, and T.~Brox, ``Deep learning
  for human part discovery in images,'' in {\em Robotics and Automation (ICRA),
  2016 IEEE International Conference on}, pp.~1634--1641, IEEE, 2016.

\bibitem{newell2016stacked}
A.~Newell, K.~Yang, and J.~Deng, ``Stacked hourglass networks for human pose
  estimation,'' in {\em European Conference on Computer Vision}, pp.~483--499,
  Springer, 2016.

\bibitem{wei2016convolutional}
S.-E. Wei, V.~Ramakrishna, T.~Kanade, and Y.~Sheikh, ``Convolutional pose
  machines,'' in {\em Proceedings of the IEEE Conference on Computer Vision and
  Pattern Recognition}, pp.~4724--4732, 2016.

\bibitem{hua2005learning}
G.~Hua, M.-H. Yang, and Y.~Wu, ``Learning to estimate human pose with data
  driven belief propagation,'' in {\em Computer Vision and Pattern Recognition,
  2005. CVPR 2005. IEEE Computer Society Conference on}, vol.~2, pp.~747--754,
  IEEE, 2005.

\bibitem{sigal2004tracking}
L.~Sigal, S.~Bhatia, S.~Roth, M.~J. Black, and M.~Isard, ``Tracking
  loose-limbed people,'' in {\em Computer Vision and Pattern Recognition, 2004.
  CVPR 2004. Proceedings of the 2004 IEEE Computer Society Conference on},
  vol.~1, pp.~I--I, IEEE, 2004.

\bibitem{yang2013articulated}
Y.~Yang and D.~Ramanan, ``Articulated human detection with flexible mixtures of
  parts,'' {\em IEEE transactions on pattern analysis and machine
  intelligence}, vol.~35, no.~12, pp.~2878--2890, 2013.

\bibitem{pishchulin2012articulated}
L.~Pishchulin, A.~Jain, M.~Andriluka, T.~Thorm{\"a}hlen, and B.~Schiele,
  ``Articulated people detection and pose estimation: Reshaping the future,''
  in {\em Computer Vision and Pattern Recognition (CVPR), 2012 IEEE Conference
  on}, pp.~3178--3185, IEEE, 2012.

\bibitem{gkioxari2014using}
G.~Gkioxari, B.~Hariharan, R.~Girshick, and J.~Malik, ``Using k-poselets for
  detecting people and localizing their keypoints,'' in {\em Proceedings of the
  IEEE Conference on Computer Vision and Pattern Recognition}, pp.~3582--3589,
  2014.

\bibitem{sun2011articulated}
M.~Sun and S.~Savarese, ``Articulated part-based model for joint object
  detection and pose estimation,'' 2011.

\bibitem{papandreou2017towards}
G.~Papandreou, T.~Zhu, N.~Kanazawa, A.~Toshev, J.~Tompson, C.~Bregler, and
  K.~Murphy, ``Towards accurate multi-person pose estimation in the wild,'' in
  {\em CVPR}, vol.~3, p.~6, 2017.

\bibitem{toshev2014deeppose}
A.~Toshev and C.~Szegedy, ``Deeppose: Human pose estimation via deep neural
  networks,'' in {\em Proceedings of the IEEE conference on computer vision and
  pattern recognition}, pp.~1653--1660, 2014.

\bibitem{pfister2015flowing}
T.~Pfister, J.~Charles, and A.~Zisserman, ``Flowing convnets for human pose
  estimation in videos,'' in {\em Proceedings of the IEEE International
  Conference on Computer Vision}, pp.~1913--1921, 2015.

\bibitem{pishchulin2016deepcut}
L.~Pishchulin, E.~Insafutdinov, S.~Tang, B.~Andres, M.~Andriluka, P.~V. Gehler,
  and B.~Schiele, ``Deepcut: Joint subset partition and labeling for multi
  person pose estimation,'' in {\em Proceedings of the IEEE Conference on
  Computer Vision and Pattern Recognition}, pp.~4929--4937, 2016.

\bibitem{insafutdinov2016deepercut}
E.~Insafutdinov, L.~Pishchulin, B.~Andres, M.~Andriluka, and B.~Schiele,
  ``Deepercut: A deeper, stronger, and faster multi-person pose estimation
  model,'' in {\em European Conference on Computer Vision}, pp.~34--50,
  Springer, 2016.

\bibitem{cao2017realtime}
Z.~Cao, T.~Simon, S.-E. Wei, and Y.~Sheikh, ``Realtime multi-person 2d pose
  estimation using part affinity fields,'' in {\em CVPR}, vol.~1, p.~7, 2017.

\bibitem{papandreou2018personlab}
G.~Papandreou, T.~Zhu, L.-c. Chen, S.~Gidaris, J.~Tompson, and K.~Murphy,
  ``Personlab: Person pose estimation and instance segmentation with a
  part-based geometric embedding model,'' 2018.

\bibitem{kanellakis2017survey}
C.~Kanellakis and G.~Nikolakopoulos, ``Survey on computer vision for uavs:
  Current developments and trends,'' {\em Journal of Intelligent \& Robotic
  Systems}, vol.~87, no.~1, pp.~141--168, 2017.

\bibitem{garcia2013guidance}
G.~J. Garcia, P.~Gil, D.~Ll{\'a}cer, and F.~Torres, ``Guidance of robot arms
  using depth data from rgb-d camera,'' 2013.

\bibitem{varhegyivisual}
T.~Varhegyi, M.~Melik-Merkumians, M.~Steinegger, G.~Halmetschlager-Funek, and
  G.~Schitter, ``A visual servoing approach for a six degrees-of-freedom
  industrial robot by rgb-d sensing,''

\bibitem{levine2018learning}
S.~Levine, P.~Pastor, A.~Krizhevsky, J.~Ibarz, and D.~Quillen, ``Learning
  hand-eye coordination for robotic grasping with deep learning and large-scale
  data collection,'' {\em The International Journal of Robotics Research},
  vol.~37, no.~4-5, pp.~421--436, 2018.

\bibitem{miseikis2018multi}
J.~Miseikis, I.~Brijacak, S.~Yahyanejad, K.~Glette, O.~J. Elle, and
  J.~Torresen, ``Multi-objective convolutional neural networks for robot
  localisation and 3d position estimation in 2d camera images,'' {\em arXiv
  preprint arXiv:1804.03005}, 2018.

\bibitem{miseikis2018transfer}
J.~Miseikis, I.~Brijacak, S.~Yahyanejad, K.~Glette, O.~J. Elle, and
  J.~Torresen, ``Transfer learning for unseen robot detection and joint
  estimation on a multi-objective convolutional neural network,'' {\em arXiv
  preprint arXiv:1805.11849}, 2018.

\bibitem{mueller2018hands}
F.~Mueller, F.~Bernard, O.~Sotnychenko, D.~Mehta, S.~Sridhar, D.~Casas, and
  C.~Theobalt, ``Ganerated hands for real-time {3D} hand tracking from
  monocular {RGB},'' in {\em {IEEE} Conference on Computer Vision and Pattern
  Recognition ({CVPR})}, pp.~49--59, 2018.

\bibitem{Johnson2017driving}
M.~Johnson{-}Roberson, C.~Barto, R.~Mehta, S.~N. Sridhar, K.~Rosaen, and
  R.~Vasudevan, ``Driving in the matrix: Can virtual worlds replace
  human-generated annotations for real world tasks?,'' in {\em {IEEE}
  International Conference on Robotics and Automation ({ICRA})}, pp.~746--753,
  2017.

\bibitem{elbasiony2018deep}
R.~Elbasiony, W.~Gomaa, and T.~Ogata, ``Deep 3d pose dictionary: 3d human pose
  estimation from single rgb image using deep convolutional neural network,''
  in {\em International Conference on Artificial Neural Networks},
  pp.~310--320, Springer, 2018.

\bibitem{tobin2017domain}
J.~Tobin, R.~Fong, A.~Ray, J.~Schneider, W.~Zaremba, and P.~Abbeel, ``Domain
  randomization for transferring deep neural networks from simulation to the
  real world,'' in {\em Intelligent Robots and Systems (IROS), 2017 IEEE/RSJ
  International Conference on}, pp.~23--30, IEEE, 2017.

\bibitem{romera2016recurrent}
B.~Romera-Paredes and P.~H.~S. Torr, ``Recurrent instance segmentation,'' in
  {\em European Conference on Computer Vision}, pp.~312--329, Springer, 2016.

\bibitem{bretzner1998feature}
L.~Bretzner and T.~Lindeberg, ``Feature tracking with automatic selection of
  spatial scales,'' {\em Computer Vision and Image Understanding}, vol.~71,
  no.~3, pp.~385--392, 1998.

\bibitem{simonyan2014very}
K.~Simonyan and A.~Zisserman, ``Very deep convolutional networks for
  large-scale image recognition,'' {\em arXiv preprint arXiv:1409.1556}, 2014.

\bibitem{hartley2003multiple}
R.~Hartley and A.~Zisserman, {\em Multiple view geometry in computer vision}.
\newblock Cambridge university press, 2003.

\bibitem{ballas2015delving}
N.~Ballas, L.~Yao, C.~Pal, and A.~Courville, ``Delving deeper into
  convolutional networks for learning video representations,'' {\em arXiv
  preprint arXiv:1511.06432}, 2015.

\bibitem{ronneberger2015u}
O.~Ronneberger, P.~Fischer, and T.~Brox, ``U-net: Convolutional networks for
  biomedical image segmentation,'' in {\em International Conference on Medical
  image computing and computer-assisted intervention}, pp.~234--241, Springer,
  2015.

\bibitem{kuhn1955hungarian}
H.~W. Kuhn, ``The hungarian method for the assignment problem,'' {\em Naval
  research logistics quarterly}, vol.~2, no.~1-2, pp.~83--97, 1955.

\bibitem{krahenbuhl2013parameter}
P.~Kr{\"a}henb{\"u}hl and V.~Koltun, ``Parameter learning and convergent
  inference for dense random fields,'' in {\em International Conference on
  Machine Learning}, pp.~513--521, 2013.

\bibitem{kingma2014adam}
D.~P. Kingma and J.~Ba, ``Adam: A method for stochastic optimization,'' {\em
  arXiv preprint arXiv:1412.6980}, 2014.

\bibitem{everingham2010pascal}
M.~Everingham, L.~Van~Gool, C.~K. Williams, J.~Winn, and A.~Zisserman, ``The
  pascal visual object classes (voc) challenge,'' {\em International journal of
  computer vision}, vol.~88, no.~2, pp.~303--338, 2010.

\end{thebibliography}
}

\end{document}